\documentclass{article}
\usepackage[utf8]{inputenc}

\usepackage{microtype}
\usepackage{graphicx}
\usepackage{subfigure}
\usepackage{booktabs} 

\usepackage[american]{babel}
\usepackage{hyperref}
\addto\extrasamerican{
}

\usepackage[accepted]{icml2021}

\icmltitlerunning{RECol: Reconstruction Error Columns for Outlier Detection}

\begin{document}

\twocolumn[
\icmltitle{RECol: Reconstruction Error Columns for Outlier Detection}

\icmlsetsymbol{equal}{*}

\begin{icmlauthorlist}
\icmlauthor{Jörn Hees}{dfki}
\icmlauthor{Dayananda Herurkar}{dfki}
\icmlauthor{Mario Meier}{bbk}
\end{icmlauthorlist}

\icmlaffiliation{dfki}{German Research Center for Artificial Intelligence (DFKI), Germany}
\icmlaffiliation{bbk}{Deutsche Bundesbank, Germany. Disclaimer: This paper represents the authors’ personal opinions and does not necessarily reflect the views of the Deutsche Bundesbank, the Eurosystem or their staff}

\icmlcorrespondingauthor{Dayananda Herurkar}{dayananda.herurkar@dfki.de}

\icmlkeywords{Machine Learning, Outlier Detection, Anomaly Detection, Pre-processing, RECol}

\vskip 0.3in
]



\makeatletter
\renewcommand{\ICML@appearing}{\vspace{-2em}}
\makeatother
\printAffiliationsAndNotice{}  

\begin{abstract}
Detecting outliers or anomalies is a common data analysis task.
As a sub-field of unsupervised machine learning, a large variety of approaches exist, but the vast majority treats the input features as independent and often fails to recognize even simple (linear) relationships in the input feature space.
Hence, we introduce RECol, a generic data pre-processing approach to generate additional columns in a leave-one-out-fashion: For each column, we try to predict its values based on the other columns, generating reconstruction error columns.
We run experiments across a large variety of common baseline approaches and benchmark datasets with and without our RECol pre-processing method and show that the generated reconstruction error feature space generally seems to support common outlier detection methods and often considerably improves their ROC-AUC and PR-AUC values.
\end{abstract}

\section{Introduction}
Detecting outliers in data is an often occurring exercise in various domains.
For example, the identification of certain diseases of patients or fraudulent behavior of individuals can often be supported by applying outlier detection algorithms to data.
Many different algorithms can be found in the literature to identify these outliers in unsupervised settings where no label is available.

However, the existing approaches apply a certain definition of an outlier that might not be appropriate for detecting outliers in all relevant contexts.
Using standard algorithms, outliers are typically defined as data points which are far apart from other data points in a global or local sense.
However, one might actually be interested in data points that deviate from underlying relations in the data.
While such deviations might coincide with points far apart from others, this must not be always the case.

\begin{figure}[t]
    \centering
    \includegraphics[trim=20px 5px 35px 30px,clip,width=\columnwidth]{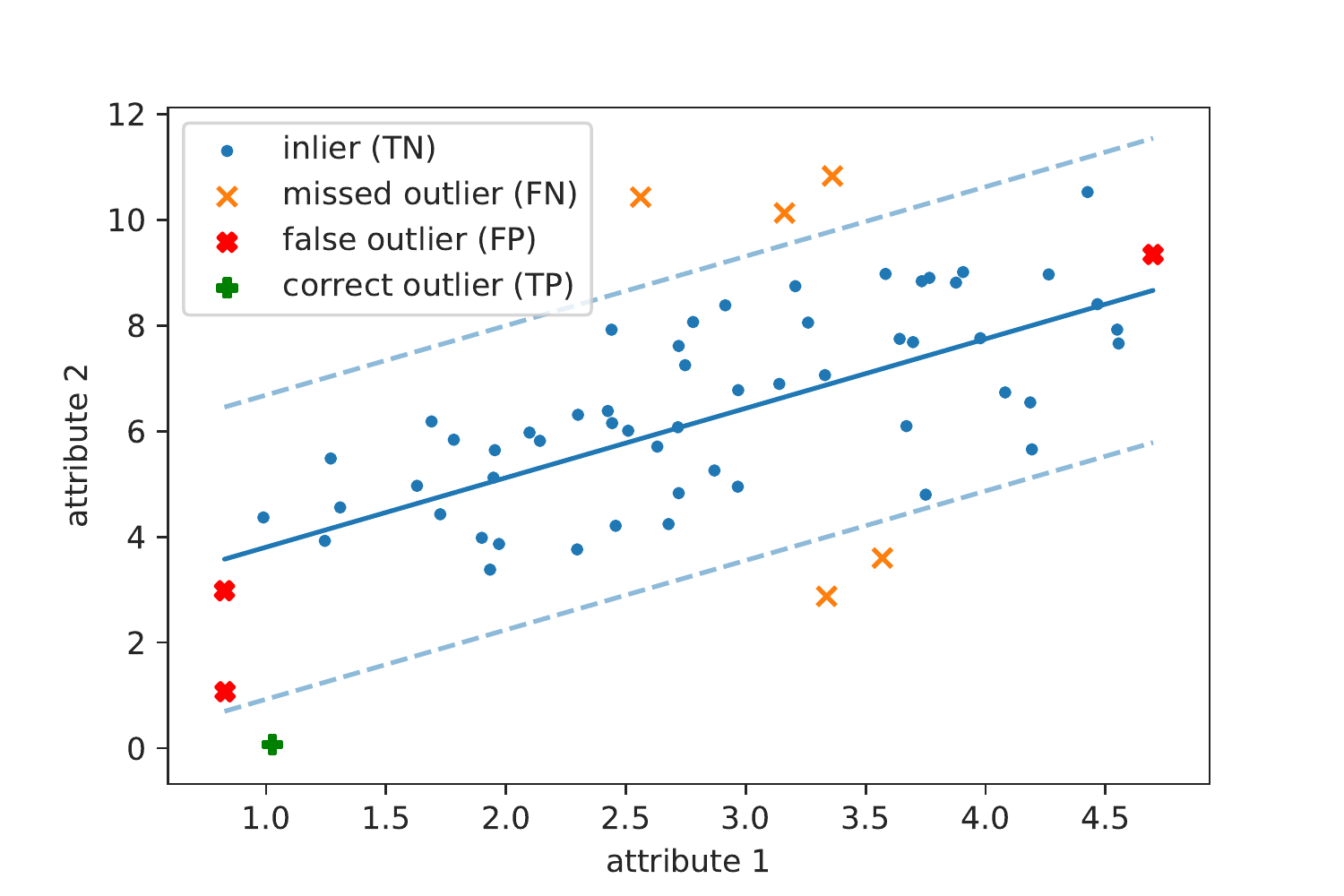}
    \caption{Example dataset with a linear relation between 2 attributes.
    The dashed lines show the 2 $\sigma$ interval.
    We can see that traditional OD methods (an isolation forest in this case) incorrectly classify points at the far ends as outliers, while missing outliers closer to the center.
    }
    \label{fig:re_illustration}
\end{figure}
We illustrate this issue in \autoref{fig:re_illustration} with a toy example in two dimensions.
The points in the scatter plot constitute some data.
Applying a common outlier detection algorithm to this data flags the red and green points as outliers.
However, notice that in our example there is a simple linear relationship between the two attributes.
Hence, we argue that an outlier could also be defined by points deviating markedly from the underlying (often latent) relationship in the data.
In our example these are the points outside the $2 \sigma$ interval of the linear relationship.
As one can see, traditional approaches often flag points inside such a bandwidth as outliers and miss points outside of it.

To tackle this problem, we present an approach to map the data to a reconstruction error space and apply the outlier detection algorithm to a combination of the previous and this novel space.
We do so by estimating a supervised model for each input feature using all other features as inputs.
Consequently, for each input feature column in the original dataset, one receives an additional (derived) feature column containing the reconstruction errors.
We can then use these new features as (additional) inputs for existing outlier detection algorithms.
This enables the algorithms to easily identify data points that can hardly be reconstructed by the relationships in the data.
In \autoref{fig:re_illustration} these would be the green and orange points, because their distance to the bandwidth is larger and therefore their reconstruction error is higher. 
In reconstruction error space they lie far apart from the other data points, making them suspicious to the outlier detection algorithm.

Therefore, we claim that by calculating additional reconstruction error columns (RECols) for each feature in the dataset, one can improve the detection of outliers in unsupervised contexts. Despite the simple example, one should note that our approach also can also be beneficial in case of more complex and non-linear relationships within the data: For the generation of RECols we can use any kind of supervised machine learning approach.

The remainder of this paper is structured as follows.
After presenting related work in \autoref{sec:relwork}, we will provide an in-depth description of our approach in \autoref{sec:approach} and evaluate our approach in \autoref{sec:evaluation} by comparing it to a variety of common approaches and benchmark datasets as described in \autoref{sec:experimentalsetup}. \autoref{sec:results} then contains our findings and a discussion, before concluding this paper in \autoref{sec:conclusion}.

\section{Related Work}\label{sec:relwork}
In the field of anomaly or outlier detection, many interesting approaches exist that try to detect irregularities of different kinds.
Good overviews over the field can be found in \cite{8130440,goldstein} and more recent in \cite{DBLP:journals/corr/abs-1901-03407}.
In this paper we do not present another outlier detection algorithm.
Instead, we present a general pre-processing idea and argue that our idea is beneficial across use-cases and existing approaches.
Hence, we base our work on Goldstein et al \cite{goldstein}.
They conduct a comprehensive study of unsupervised anomaly detection algorithms for multivariate data, in which they analyse 19 different unsupervised anomaly detection algorithms (cluster, distance, density, and statistics based) using 10 different standard datasets, resulting in a detailed analysis of the advantages and disadvantages of each algorithm.
In this paper, we hence treat their models and results as baseline results and compare our approach with these baselines to analyze the advantages generated by our pre-processing approach for unsupervised anomaly detection in general.

Our approach is related to other techniques such as principle component analysis (PCA) or auto-encoders (AEs).
PCA and AEs are often used as dimensionality reduction techniques.
Errors generated by a reduced set of principle components of a PCA during the reconstruction of the original data have been used for outlier detection \cite{7098354, CHEN20093706}.
Analogously, there are many approaches for outlier detection based on the reconstruction errors of AEs \cite{Sakurada, Zhou, Xia, 2019arXiv190402639G, 10.1371/journal.pone.0225991}.
In contrast to PCA and AEs, our idea is not motivated by a dimensionality reduction or compression, but we instead (in general) increase the number of dimensions.
Additionally, our approach results in one reconstruction error component per column for each sample and not just one (often difficult to decompose or to correctly attribute to its source) reconstruction error for the whole sample.
Finally, our approach is not intended as a direct outlier detection algorithm, but as an enrichment of the input-space.
In other words, our approach is intended as a pre-processing step for an arbitrary outlier detection algorithm that might follow.

\section{Approach}\label{sec:approach}
\begin{figure*}
    \centering
    \includegraphics[height=0.8\columnwidth, width=2.1\columnwidth]{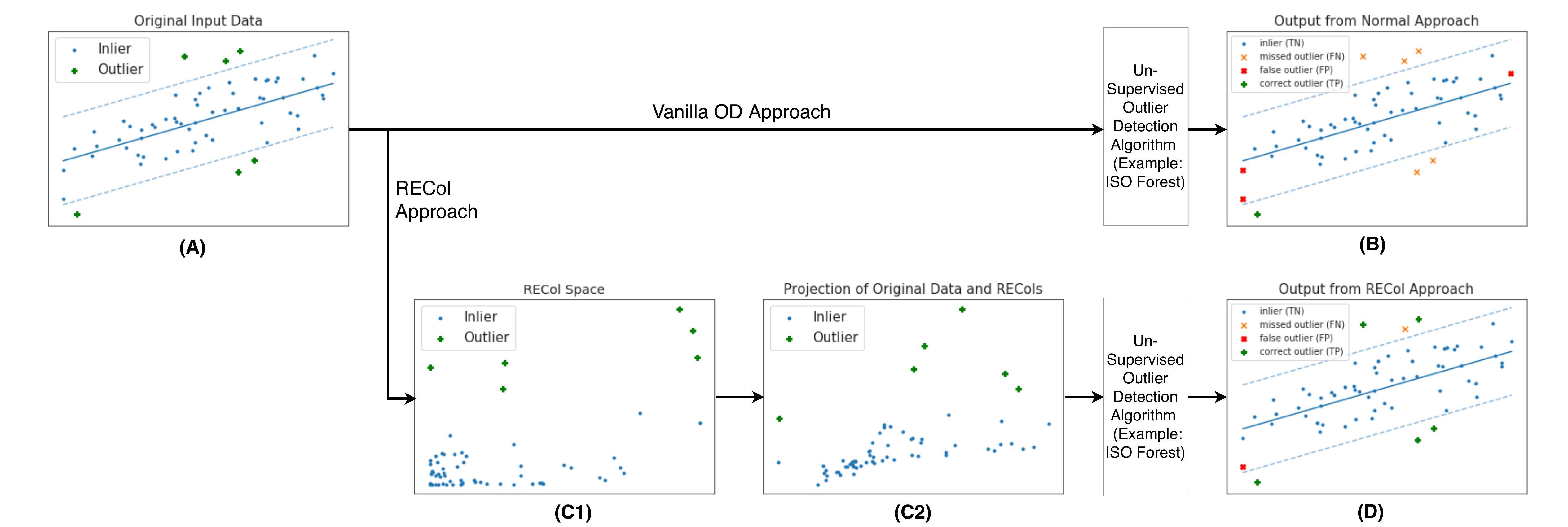}
    \vspace{-1em}
    \caption{Reconstruction Error Columns Approach.
    (A) Input data in 2D space with blue points as inliers and green points as outliers.
    (B) Result of directly applying a standard outlier detection algorithm to the input data missing obvious relations between attributes.
    (C1) The 2D Reconstruction Error space.
    (C2) A 2D projection of the 4D concatenation of the original and RECol space.
    We can see that in (C1) and (C2) the true outliers are easier separable from the inliers.
    (D) Results of the same standard outlier detection algorithm using our (RECol) pre-processing approach, resulting in a better prediction of true outliers and less misclassifications of outliers compared to (B).
    }
    \label{fig:re_intuition}
\end{figure*}

The standard approach for detecting outliers in unsupervised settings follows this outline: (1) pre-processing, (2) outlier detection algorithm and (3) result evaluation.
We extend this approach by expanding step (1):
For each column in the dataset we use a supervised machine learning approach to predict this column based on all other features in the dataset (so excluding the one to be predicted, hence also denoted as a leave-one-out approach).
After training of the supervised model for a specific column we calculate the reconstruction (or prediction) error for each data point (e.g., the mean squared error or some other metric of the prediction error) in the train and test set.
Repeating this process, we arrive at a corresponding reconstruction error feature column (RECol) for each of the original columns, effectively doubling the number of features.
After this adjusted first step, we can continue with (2), the outlier detection algorithm.
However, one might consider to also pre-process the RECols first, e.g. scaling them appropriately. 

\autoref{fig:re_intuition} illustrates our extended approach for outlier detection. In \autoref{fig:re_intuition} (A) we plot a dataset with two features, where outliers are marked in green and inliers in blue. In this dataset an outlier is a point that markedly deviates from the underlying relationship between the two features. By directly applying an outlier detection algorithm to the data one usually receives predictions as highlighted by the red and green points in \autoref{fig:re_intuition} (B), missing the orange outliers and delivering a dissatisfying performance. This is caused by the outlier detection algorithms focusing on labeling data points that are far away from others in the data space.
By applying our approach where we create RECols and apply the outlier detection algorithm to these features, we are able to classify points as outliers that deviate from the underlying relationship, as can be seen in \autoref{fig:re_intuition} (C1) and (C2). (C1) shows the 2D reconstruction error space and (C2) a 2D projection of the 4D space arising from a concatenation of the original data space with the RECols. One can see, that the true outliers are easier to separate in these new spaces introduced by our approach. Given these new representations, the following the outlier detection model is better able to classify outliers correctly, as can be seen in \autoref{fig:re_intuition} (D).

\subsection{Possible Design Decisions}\label{sec:approachQuestions}
Aside from whether our approach leads to improvements in detecting outliers in data, our approach opens up a few questions regarding the specification of the input space for the outlier detection algorithm:

(a) Should one use only the RECols for estimation or the combined feature space with the original features and additional RECols? If one is only concerned about outliers that are hard to reconstruct from the underlying relationship in the data one can in principle estimate the outlier detection model abolishing the initial data features. However, they might still contain valuable information for estimation. Therefore, in our evaluation we compare the results of the outlier detection algorithm on the initial data features only, the RECols only and the combined (concatenated) feature space.

(b) Can the RECols be used as an outlier detection algorithm directly by fusing the reconstruction errors appropriately? In principle, one could directly use the RECols and aggregate them to calculate an outlier score for each data point. To test whether this works we normalize the RECols and aggregate them to an outlier score by taking the average reconstruction error across columns. In the next section we compare our results from this exercise with standard outlier detection algorithms. 

(c) How should the supervised algorithm to calculate RECols be specified? Ideally, we want to estimate the true hypothesis in the data. In practice, however, there is noise in the data and dataset is incomplete in the sense that relevant features are missing in the data. Therefore, we want to estimate a good approximation to the true underlying relationship. In our experiments we use a wide range of algorithms to check whether the choice of the supervised model affects the results.

(d) Finally, how should RECols be calculated and pre-processed and should all newly created columns be used for estimating the outlier detection algorithm? That is, which reconstruction error metric should be applied, e.g. mean squared error, mean absolute deviation, or some other prediction error metric? In terms of pre-processing one can think about scaling the attributes, dropping uninformative RECols or clipping RECol values.

To evaluate whether our RECol approach is beneficial and to answer these additional questions, we have run several experiments, as described in the following.

\section{Evaluation}\label{sec:evaluation}

\subsection{Experimental Setup}\label{sec:experimentalsetup}
In this section we evaluate whether using RECols leads to improvements when looking for outliers in data.
We measure the performance of an algorithm by either the area under the receiver operating characteristic (ROC-AUC) or the area under the precision recall curve (PR-AUC), which are standard metrics in the outlier detection domain.
Before estimating RECols or applying an outlier detection model we split the dataset to a train set containing 70\% of the data and a test set with the remaining 30\% of data.
Then we estimate the RECols and outlier detection model on the train set and evaluate them on the test set to avoid data dredging.
We calculate RECols in the test data using the trained model and evaluate the outlier detection model with these data.
Finally we perform three types of experiments: (a) Estimating baseline results using only normal data features, (b) using only RECols for estimation and (c) using both types of columns combined (concatenated) for estimating the outlier detection model.

\subsubsection{Baselines}
We compare our results to those of \cite{goldstein} using the same set of datasets and the most common outlier detection algorithms as in their paper. Similar to \cite{goldstein} we optimize hyper-parameters using a grid search to reproduce their results. In the end we find very similar results and pick the best outlier detection model for each algorithm and each dataset to compare to our RECol approach. We apply the following algorithms: HBOS \cite{hbos}, ISO \cite{iso}, KNN \cite{kthnn}, KthNN \cite{kthnn}, LDCOF \cite{ldcof}, LOF \cite{lof}, LOoP \cite{loop}, Nu-OCSVM \cite{uocsvm}, OCSVM \cite{ocsvm}, CBLOF \cite{cblof}, uCBLOF \cite{cblof}. These are typical outlier detection algorithms from the literature. We refer the reader to \cite{goldstein} for a more detailed discussion of the different algorithms. 

In addition to the aforementioned baseline results, we initially also experimented with several auto-encoder based approaches for outlier detection, such as \cite{Sakurada, Zhou, Xia, 2019arXiv190402639G, 10.1371/journal.pone.0225991}.
However, despite experimenting with different architectures, varying the number of hidden layers and the size of the smallest compression layer we found that our auto-encoder based attempts were only ever among the best models (with negligible improvement) on the kdd99 dataset.
We reason that this is likely due to the large amount of training data necessary for such approaches.
Due to this and the large number of additional design decisions, we decided to leave an in-depth analysis of a combination of auto-encoders with our RECol approach on multiple larger datasets for future work.

\subsubsection{Datasets}
As in \cite{goldstein} we apply all of our experiments to the following ten datasets: aloi, annthyroid, breast-cancer, kdd99, letter, pen-global, pen-local, satellite, shuttle and speech \cite{Bache+Lichman:2013, Micenkov2014LearningOE, aloi, goldstein}. The datasets are very distinct, in particular in their number of observations, their dimensionality and their fraction of outliers in the dataset. This allows us to analyse the effect of RECols on a wide range of different types of datasets. A more detailed description of the datasets itself can be found in \cite{goldstein}.

\subsubsection{Parameters}
To give an answer to question (a) whether one should use RECols for outlier detection we decided to run many different experiments.
By applying different experiments we can also learn about question (c) how one should specify the supervised algorithm for our leave-one-out approach.
Finally, to answer (d) we need to try out different approaches to generate and process RECols, like the distance metric and whether all RECol columns should be used for training the outlier detection model.

Therefore, we experiment with various specifications of training and generating the RECols.
For each fixed best parameter baseline OD model and dataset, we experiment with the following parameters for the RECol models:
\begin{enumerate}
    \item Supervised algorithms: Decision Tree Regressor, Random Forest Regressor, Gradient Boosting Regressor, Linear Regression, Support Vector Regression with RBF kernel
    \item Reconstruction error calculation: Mean Squared Error, Mean Absolute Deviation
    \item Scaling of input features for supervised algorithm: Min-max Scaler, Standard Scaler
    \item Clipping reconstruction errors at twice the standard deviation of the RECol
    \item Using only subsets of RECols depending on the $R^2$ metric of the RECol model: dropping RECols with $R^2$ values below ($0.05$, $0.10$) or above ($0.95$, $0.90$) certain thresholds
\end{enumerate}
In total, we evaluated 895 different ways of creating RECols for each fixed best parameter baseline OD model and dataset.
We evaluated them against the corresponding fixed best parameter baseline OD model for all datasets, expect kdd99.
For kdd99 we ran only eight experiments due to the size of the dataset and the time each experiment consumes.\footnote{This can only harm us, because we might miss many possible experiments that could outperform the baseline.}
We will discuss our results in the following subsection.

\subsection{Results \& Discussion}\label{sec:results}

\begin{table}\centering
\caption{RECols versus Baseline Results with ROC-AUC}
\begin{tabular}{lrrr}
\toprule
      Dataset  &  Baseline &  RECols &    $\Delta$ \\
\midrule
     pen-local &          99.47 &   99.65 &        0.18 \\
    pen-global &          98.89 &   98.11 &       -0.78 \\
 breast-cancer &         100.00 &   98.46 &       -1.54 \\
        speech &          66.14 &   61.79 &       -4.35 \\
          aloi &          78.52 &   80.37 &        1.85 \\
       shuttle &          99.85 &   99.94 &        0.09 \\
        letter &          89.50 &   92.78 &        3.28 \\
     satellite &          96.90 &   95.51 &       -1.39 \\
    annthyroid &          72.09 &   89.58 &       17.49 \\
         kdd99 &          99.96 &   99.83 &       -0.13 \\
\bottomrule
\multicolumn{4}{l}{\begin{minipage}{\columnwidth-2\tabcolsep}
\vspace{1ex}\small\textit{Notes:} Performance of OD-model is measured in ROC-AUC metric. Baseline is the best OD-Algorithm using standard features only. RECols is the best algorithm adding RECols in addition.
\end{minipage}}
\end{tabular}
\label{tab:recols}
\end{table}

\begin{figure*}[p]
    \centering
    \includegraphics[width=.39\textwidth]{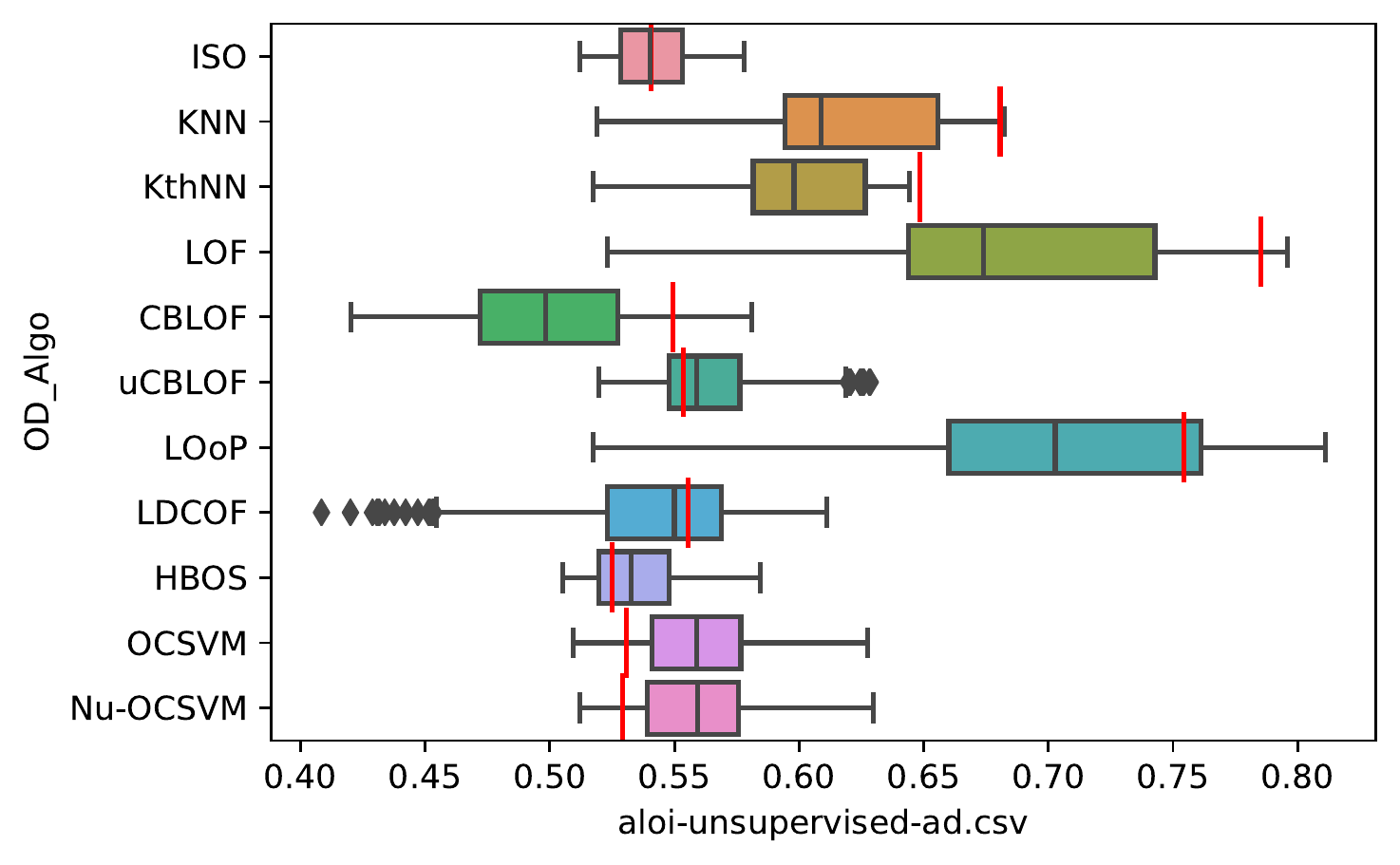}
    \hspace{3em}
    \includegraphics[width=.39\textwidth]{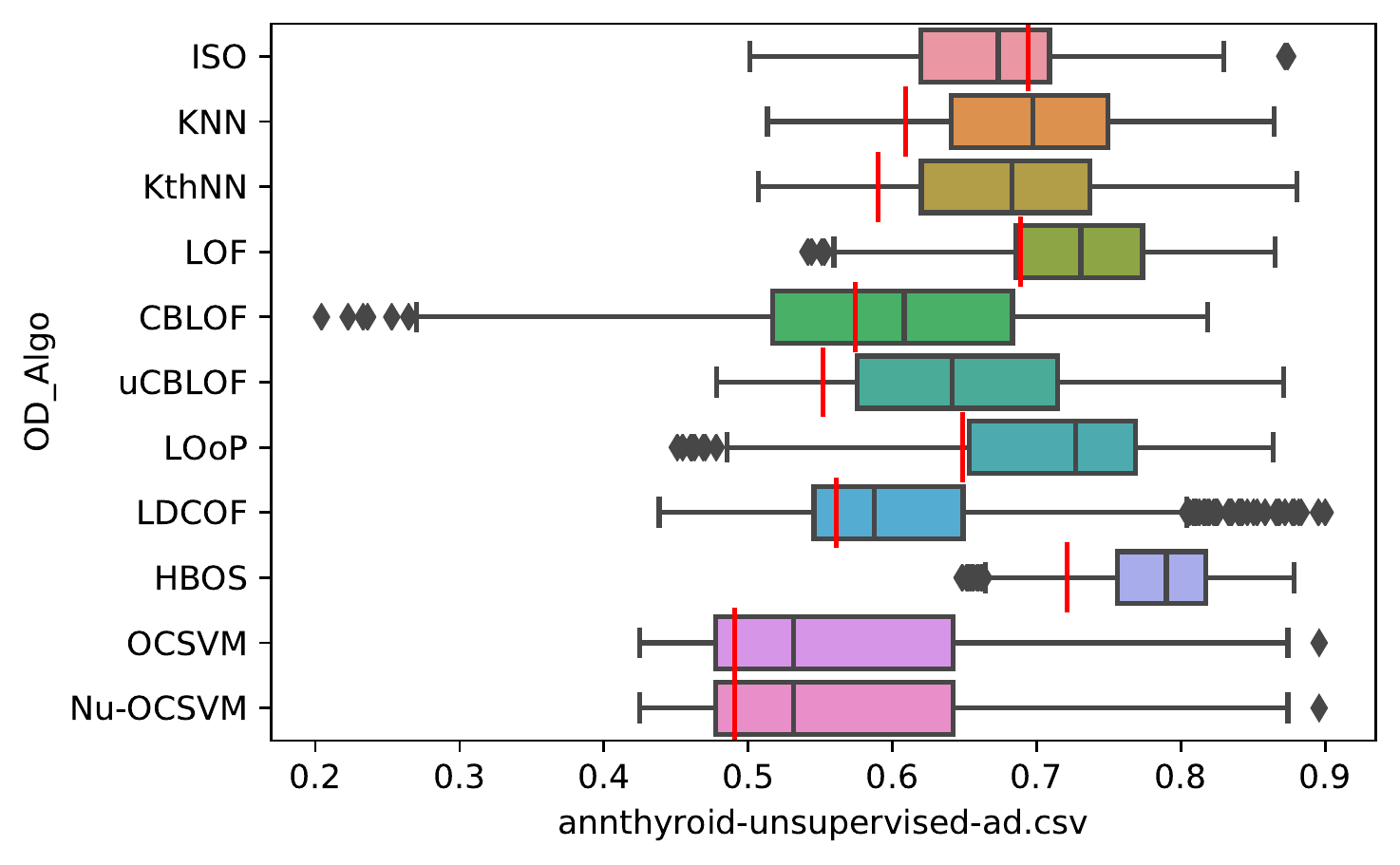}
    
    \includegraphics[width=.39\textwidth]{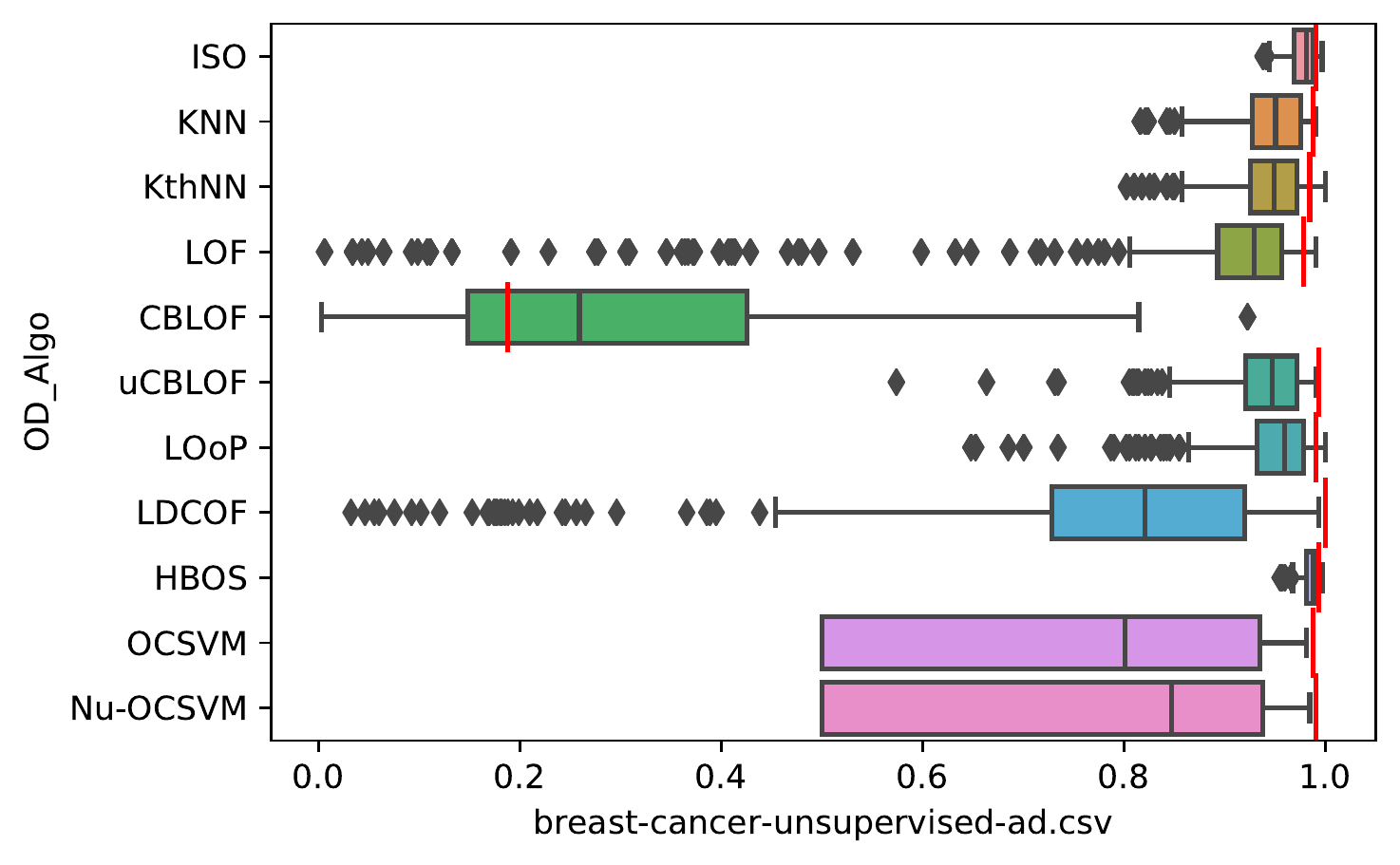}
    \hspace{3em}
    \includegraphics[width=.39\textwidth]{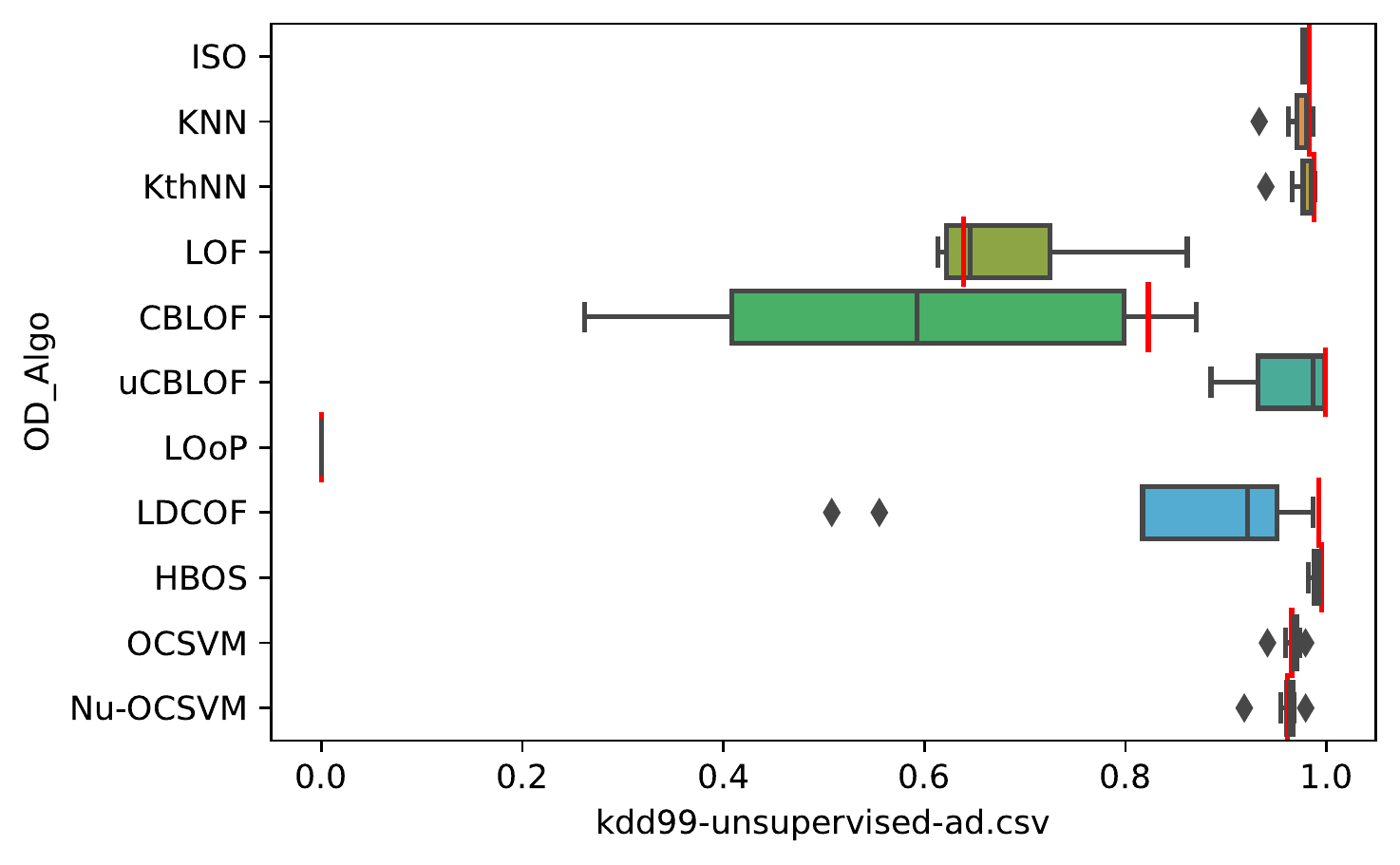}
    
    \includegraphics[width=.39\textwidth]{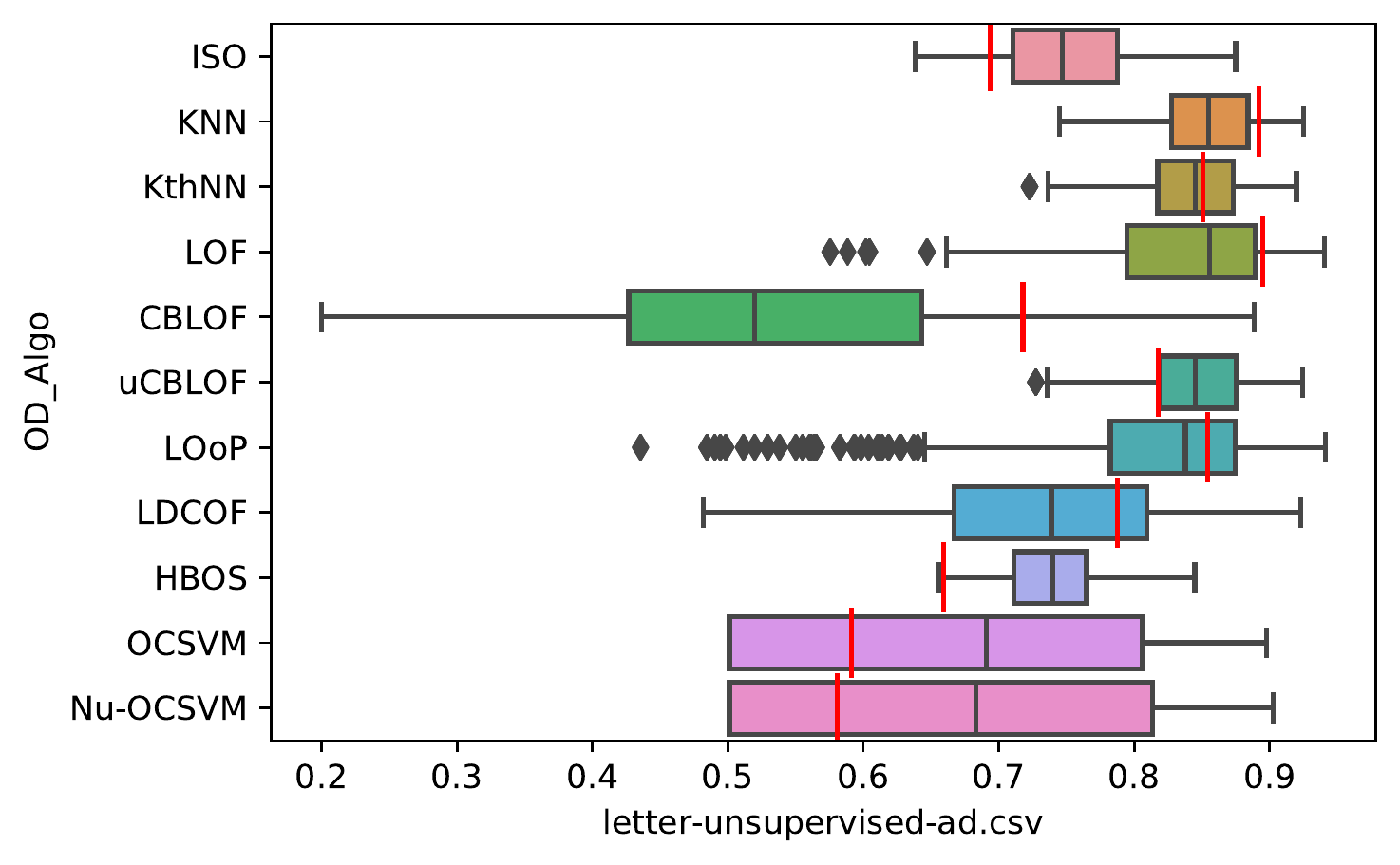}
    \hspace{3em}
    \includegraphics[width=.39\textwidth]{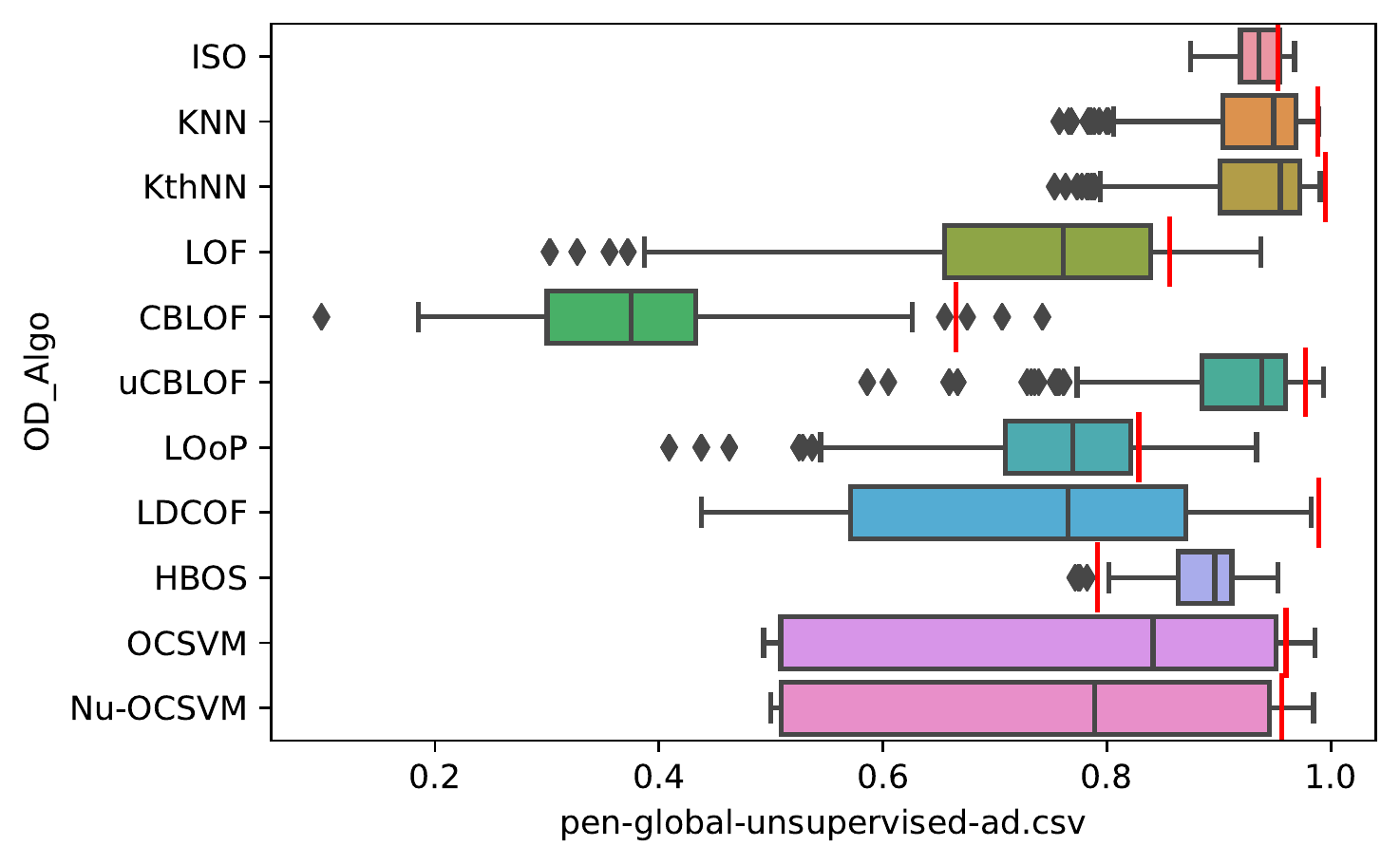}
    
    \includegraphics[width=.39\textwidth]{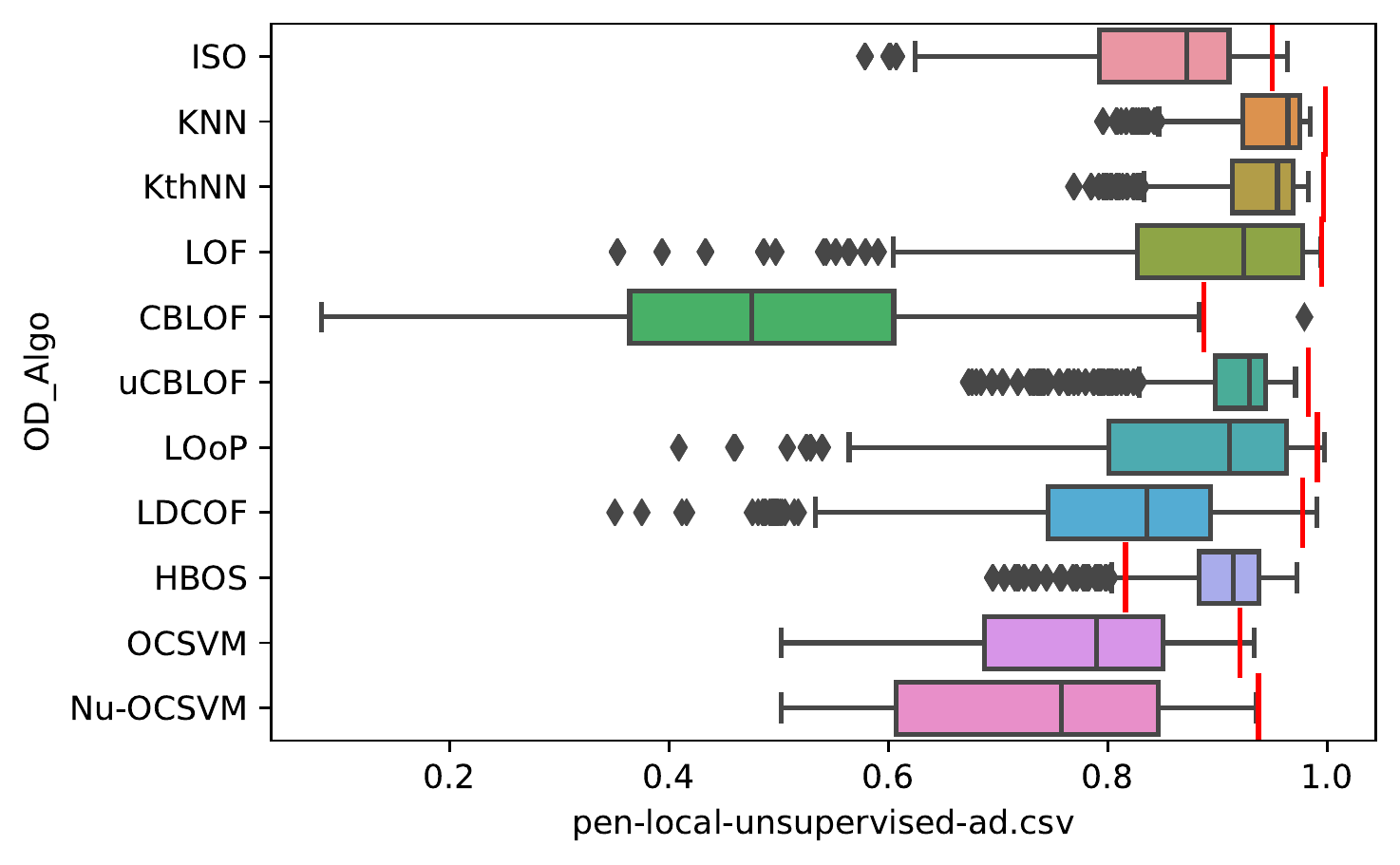}
    \hspace{3em}
    \includegraphics[width=.39\textwidth]{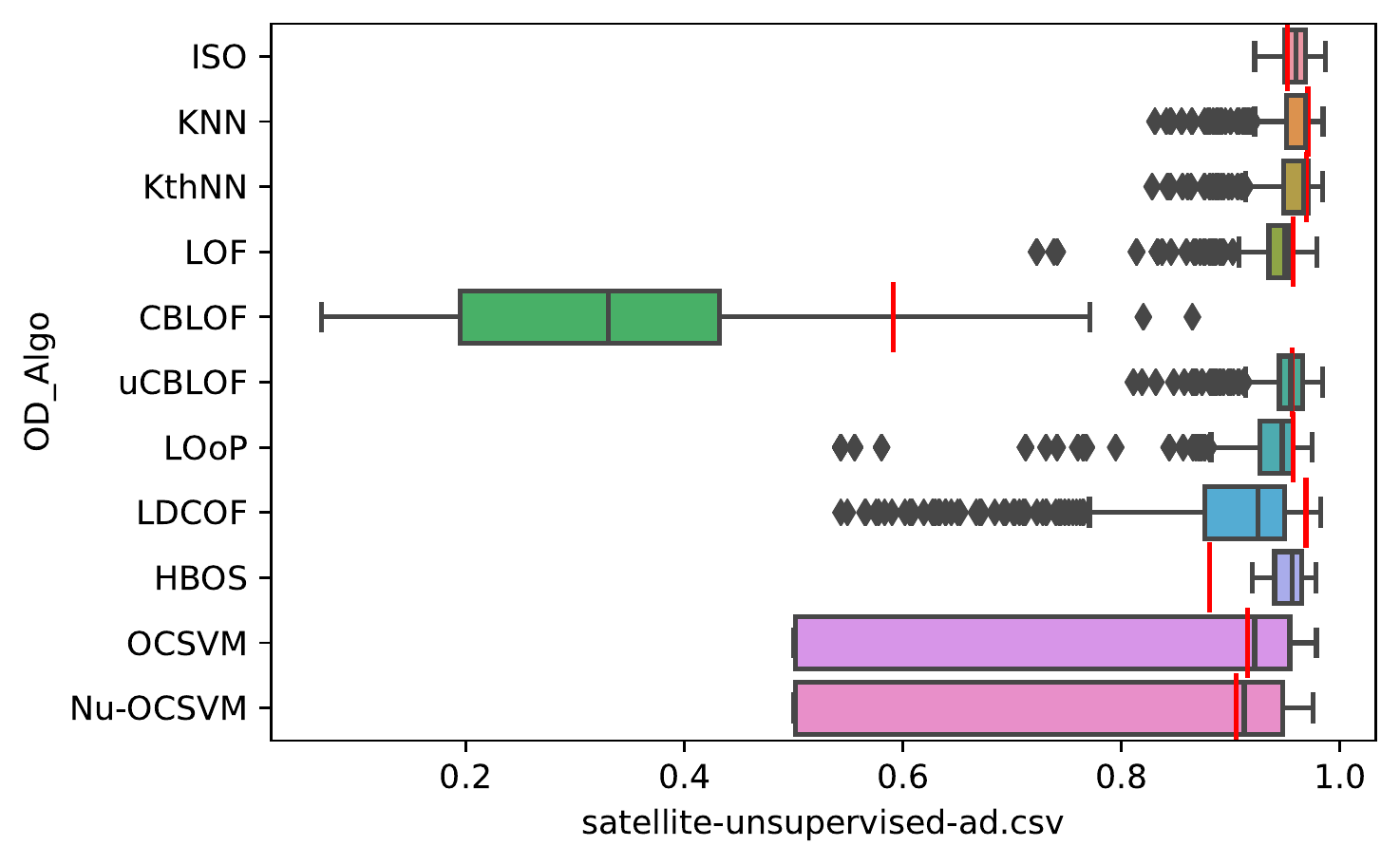}
    
    \includegraphics[width=.39\textwidth]{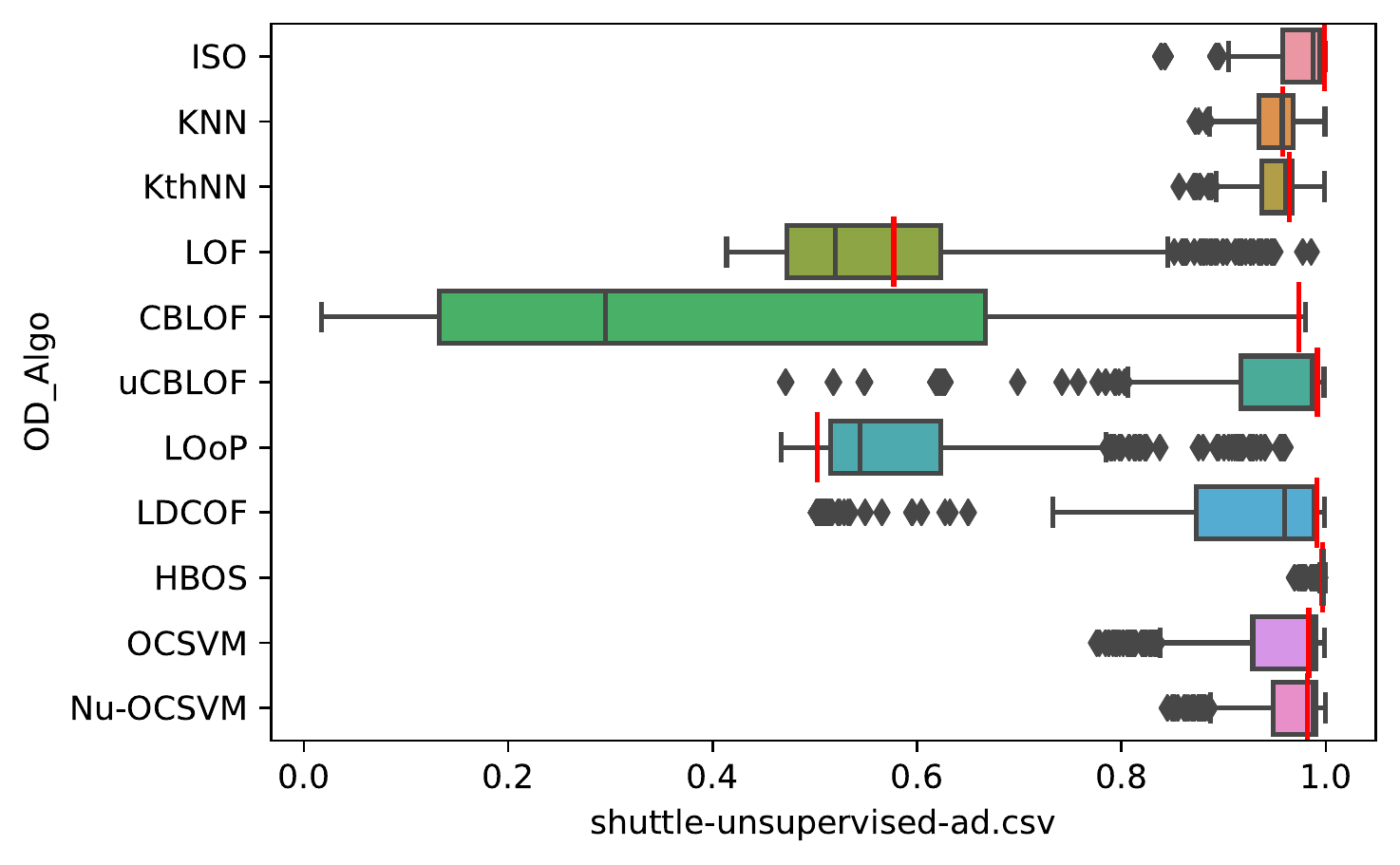}
    \hspace{3em}
    \includegraphics[width=.39\textwidth]{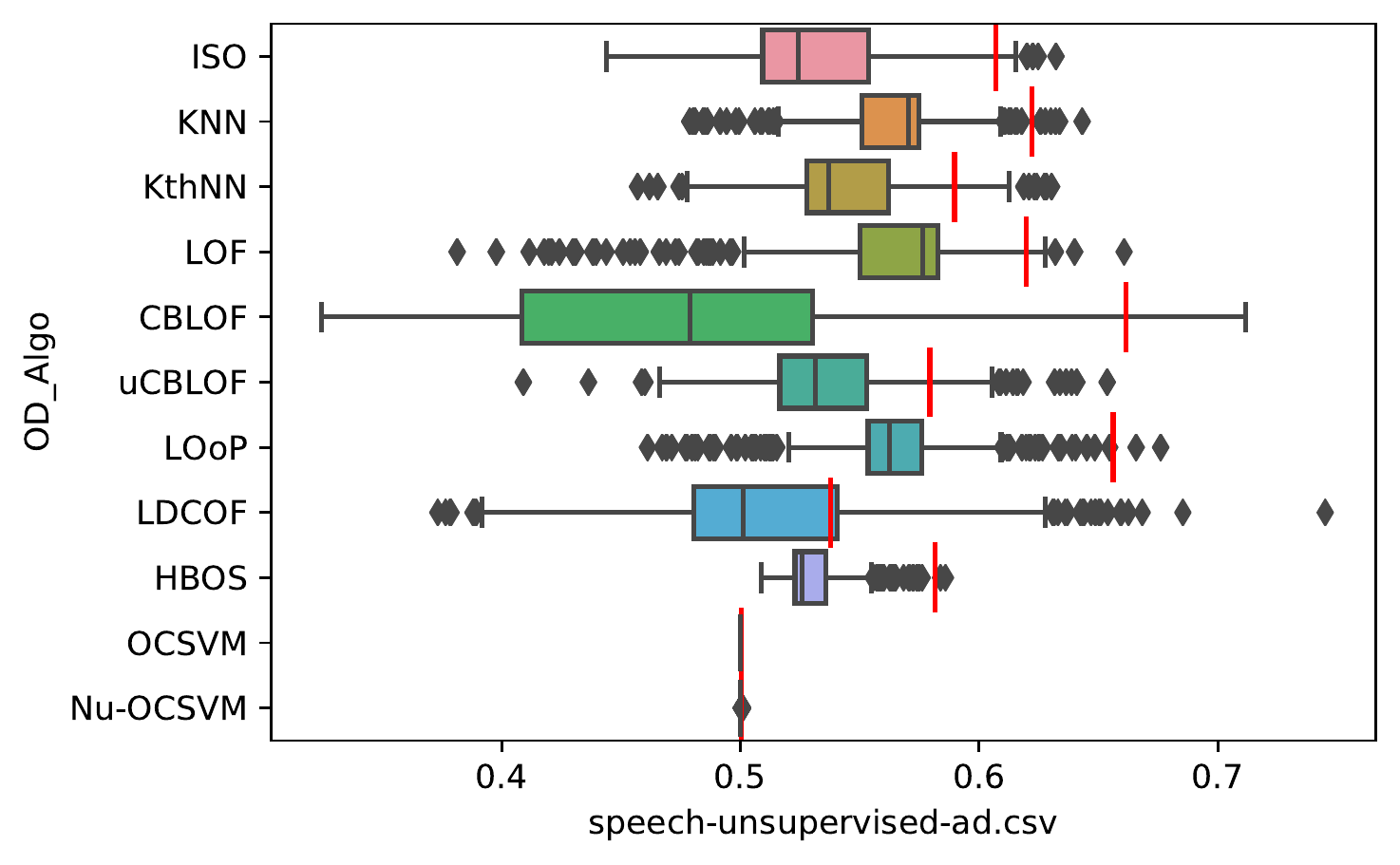}
    \vspace{-1ex}
    \caption{Boxplot results by algorithm and data set. Each sub-figure shows the results for one dataset. The distributions of ROC-AUC are shown for our RECol approach based on each of the underlying OD algorithms. The red vertical line shows the best (maximum) baseline result (without RECol). By this we give the baselines the advantage of already picking the best parameters, while showing ROC-AUC values across 895 such parameters for the RECol approaches.}
    \label{fig:boxplots}
\end{figure*}

To evaluate whether RECols help to better detect outliers in data we make the following comparison and start with the ROC-AUC metric for evaluation.
We define the best baseline model as the maximum ROC-AUC value across all standard outlier detection algorithms where we also optimized hyper-parameters using a grid search as in \cite{goldstein}.
The best model is picked as the best ROC-AUC model in the train dataset while we then compare their performance in the test dataset.
In \autoref{tab:recols} these values are shown in the second column.
Similarly, we take the best RECol model by picking the best train model in terms of the ROC-AUC value across our experiments.
However, we do not optimize hyper-parameters of the outlier detection model for RECol approach but take the same hyper-parameter specification as for the baseline.\footnote{These choices can only harm us in that we restrict ourselves to less options compared to the baseline results.}
The third column of \autoref{tab:recols} shows our results in the test sample.
In total, we improve in 5 out of 10 datasets.
In light of the fact that for half of the datasets the baseline is very close to 100 \% (due to the low contamination in the dataset), this is a relatively strong result.
We have also compared the best baseline against our RECol approach for each dataset and outlier detection algorithm separately.
The average improvement in this setup is around 6 percentage points in terms of ROC-AUC and RECols improve the results in over 2/3 of cases.
In over 1/3 of cases the improvement is even above 5 percentage points in the ROC-AUC metric.
When we only look at baselines with a baseline metric below 95\% ROC-AUC we see in almost 60\% of settings an improvement of over 5 percentage points in the ROC-AUC metric by just adding RECols as additional input to the outlier detection model.
A visualization of the different parameter choices for each dataset and RECol based on each of the baseline OD algorithms can be found in \autoref{fig:boxplots}.
In principle, one could further improve the results by optimizing hyper-parameters of each of the underlying OD models, after choosing the one of the 895 ways to generate RECols. However, due to combinatorial explosion of experiments, we decided to instead treat the OD models' parameters as fixed (to our disadvantage).

\begin{table}\centering
\caption{RECols versus Baseline Results with PR-AUC}
\begin{tabular}{llrrr}
\toprule
      Dataset  &  Baseline &  RECols &    $\Delta$ \\
\midrule
     pen-local &     17.57 &        50.08 &       32.51 \\
    pen-global &     93.83 &        86.99 &       -6.84 \\
 breast-cancer &     85.00 &        79.37 &       -5.63 \\
        speech &      2.24 &        50.81 &       48.57 \\
          aloi &     11.65 &         9.34 &       -2.31 \\
       shuttle &     98.31 &        96.36 &       -1.95 \\
        letter &     40.26 &        53.13 &       12.87 \\
     satellite &     55.83 &        68.07 &       12.24 \\
    annthyroid &     15.67 &        53.73 &       38.06 \\
         kdd99 &     71.17 &        69.42 &       -1.75 \\
\bottomrule
\multicolumn{4}{l}{\begin{minipage}{\columnwidth-2\tabcolsep}
\vspace{1ex}\small\textit{Notes:} Performance of OD-model is measured in PR-AUC metric. Baseline is the best OD-Algorithm using standard features only. RECols is the best algorithm adding RECols in addition.
\end{minipage}}
\end{tabular}
\label{tab:recols_pr}
\end{table}

The ROC-AUC metric, however, has its weaknesses because when the outlier contamination in the data is low, the ROC-AUC value is high by construction. In particular, many datasets in our sample have a very low contamination which leads to very high ROC-AUC values independent from the quality of the outlier detection model. The comparison across algorithms and our RECol approach is still valid, but possible improvements can be very small by construction.

Therefore we perform the exact same experiments using the area under the precision recall curve (PR-AUC) to test whether using RECols improves the results compared to standard outlier detection algorithms. The PR-AUC is more stable when the dataset is very imbalanced. The results are shown in \autoref{tab:recols_pr}. The RECol approach leads to stunning improvements in five out of ten datasets while there is no improvement in the other five datasets. The average improvement in this setup is around 8.6 percentage points in terms of PR-AUC and RECols improve the results in over 75 percent of cases. In over 55 percent of cases the improvement is even above 5 percentage points in the PR-AUC metric. Hence, there are considerable improvements in the PR-AUC metric when RECols are added to the outlier detection algorithm.\\

\begin{table}[thb]\centering
\caption{Combined (RECols+Standard) features versus RECols only Results with ROC-AUC}
\begin{tabular}{lrr}
\toprule
      Dataset  &   RECols + Standard &    RECols Only \\
\midrule
     pen-local &          99.65 &        99.65 \\
    pen-global &          \textbf{98.11} &        98.05 \\
 breast-cancer &          98.46 &        \textbf{99.69} \\
        speech &          61.79 &        \textbf{63.16} \\
          aloi &          80.37 &        \textbf{81.10} \\
       shuttle &          \textbf{99.94} &        99.65 \\
        letter &          \textbf{92.78} &        86.60 \\
     satellite &          95.51 &        \textbf{97.15} \\
    annthyroid &          \textbf{89.58} &        85.02 \\
         kdd99 &          \textbf{99.83} &        99.15 \\
\bottomrule
\multicolumn{3}{l}{\begin{minipage}{\columnwidth-2\tabcolsep}
\vspace{1ex}
\small
\textit{Notes:} Performance of OD-model is measured in ROC-AUC metric. RECols Only is the best OD-Algorithm using exclusively RECol features. RECols is the best algorithm using combination of RECols and standard features in addition.
\end{minipage}}
\end{tabular}
\label{tab:recolsvsonlyre}
\end{table}

\begin{table*}[tbh]
\caption{RECol-OD versus Baseline Results with ROC-AUC}
\centering
\begin{tabular}{lrrrrr}
\toprule
       Dataset &       Best Baseline &       Avg. Baseline &    RECol-OD &         $\Delta$ to Best &            $\Delta$ to Avg. \\
\midrule
      pen-local &               99.47 &                  95.03 &    95.93 &                    -3.54 &                        0.90 \\
     pen-global &               98.89 &                  90.54 &    94.86 &                    -4.03 &                        4.32 \\
  breast-cancer &              100.00 &                  91.69 &    98.77 &                    -1.23 &                        7.07 \\
         speech &               66.14 &                  58.69 &    57.19 &                    -8.95 &                       -1.50 \\
           aloi &               78.52 &                  60.48 &    63.39 &                   -15.13 &                        2.91 \\
        shuttle &               99.85 &                  90.16 &    99.43 &                    -0.42 &                        9.27 \\
         letter &               89.50 &                  75.82 &    87.93 &                    -1.57 &                       12.10 \\
      satellite &               96.90 &                  91.12 &    97.54 &                     0.64 &                        6.41 \\
     annthyroid &               72.09 &                  60.18 &    88.90 &                    16.81 &                       28.71 \\
          kdd99 &               99.96 &                  86.17 &    99.34 &                    -0.62 &                       13.17 \\
\bottomrule
\multicolumn{6}{l}{\begin{minipage}{.85\textwidth-2\tabcolsep}
\vspace{1ex}
\small
\textit{Notes:} The performance of the OD-models is measured in the ROC-AUC metric. The best baseline is the best OD-Algorithm using standard features only. Average baseline is the average ROC-AUC value across all standard OD-models. RECol-OD here is the direct fusion of the RECols from our approach into an outlier score.
\end{minipage}}
\end{tabular}
\label{tab:recolod}
\end{table*}

\begin{table*}[tbh]
\caption{RECol-OD versus Baseline Results with PR-AUC}
\centering
\begin{tabular}{lrrrrr}
\toprule
      Dataset &       Best Baseline &         Avg. Baseline &   RECol-OD &         $\Delta$ to Best &        $\Delta$ to Avg. \\

\midrule
     pen-local &               17.57 &                  11.66 &      2.08 &                   -15.49 &                   -9.59 \\
    pen-global &               93.83 &                  71.61 &     73.84 &                   -19.99 &                    2.24 \\
 breast-cancer &               85.00 &                  71.57 &     81.67 &                    -3.33 &                   10.09 \\
        speech &                2.24 &                  10.81 &      7.10 &                     4.86 &                   -3.72 \\
          aloi &               11.65 &                   5.98 &      9.55 &                    -2.10 &                    3.56 \\
       shuttle &               98.31 &                  62.45 &     86.87 &                   -11.44 &                   24.42 \\
        letter &               40.26 &                  19.60 &     41.96 &                     1.70 &                   22.36 \\
     satellite &               55.83 &                  49.16 &     68.89 &                    13.06 &                   19.72 \\
    annthyroid &               15.67 &                   5.98 &     34.50 &                    18.83 &                   28.53 \\
         kdd99 &               71.17 &                  31.65 &     25.91 &                   -45.26 &                   -5.74 \\
\bottomrule
\multicolumn{6}{l}{\begin{minipage}{.85\textwidth-2\tabcolsep}
\vspace{1ex}
\small
\textit{Notes:} The performance of the OD-models is measured in the PR-AUC metric. The best baseline is the best OD-Algorithm using standard features only (picked on training split, which allows for the best value in test to be lower than the average, as can be seen in the speech dataset). Average baseline is the average PR-AUC value across all standard OD-models. RECol-OD here is the direct fusion of the RECols from our approach into an outlier score.
\end{minipage}}
\end{tabular}
\label{tab:recolod_pr}
\end{table*}

\subsection{Possible Design Decisions}
We will now attempt to answer the questions raised in \autoref{sec:approachQuestions}.

(a) Should one replace standard columns with RECols or use both columns for training? In \autoref{tab:recolsvsonlyre} we compare the results of outlier detection algorithms using RECols only and a combination of RECols and standard features on different datasets. In 4 out of 10 datasets RECols only approach outperforms the combination of RECols and standard features. The results using RECols only are still promising, but overall the combination of standard attributes and RECols delivers better results. We reason that this is the case because standard attributes might still contain valuable information for outlier detection. The larger dimensionality of the space with RECols has not led to problems in our experiments.

(b) Can RECols directly be used as an outlier detection algorithm? To test this we have applied the following approach:
We normalize the RECols using a min-max scaler and take the average across all columns to receive an outlier score for each observation.
When we take this outlier score and evaluate them using the ROC-AUC metric we can compare this to other outlier detection algorithms.
The results for the ROC-AUC metric are summarized in \autoref{tab:recolod}.
We find that for 2 out of 10 datasets such a simple fusion of the RECols from our approach, called RECol-OD in the following, outperforms all other baselines.
When we compare RECol-OD to the average performance of other algorithms we see that in 9 out of 10 datasets the RECol-OD algorithm performs better.
Overall, RECol-OD outperforms many standard algorithms by a significant amount.
This is somewhat surprising in light of the simple fusion method we applied.
As before, we do the same comparison using the PR-AUC metric for evaluation as can be seen in table \autoref{tab:recolod_pr}.
In 4 out of 10 datasets the RECol-OD algorithm is the most successful algorithm.
In 7 out of 10 datasets the performance is considerably higher than the average algorithm.

(c) How should the supervised algorithm to calculate RECols be specified? One has many degrees of freedom to specify the RECol model, in particular the choice of algorithm could be important. Hence, we have tried various algorithms and found that random forests or gradient boosting are relatively robust and good benchmarks. Depending on the dataset at hand other RECol algorithms might be more beneficial though. We have also reasoned about whether the performance of the RECol model impacts the performance of the outlier detection, in particular one could think of an inverse u-shape here. This is because a RECol model with an accuracy of 100\% delivers a RECol column full of zeros while a RECol model with an accuracy of 0\% delivers the square of the feature to be predicted (in case of a mean squared error RECol). In both cases adding RECols should not improve outlier detection. Hence a model that is good but not perfect might be ideal. We have checked and indeed one finds a slight inverse u-shape relation between the RECol accuracy and the outlier detection ROC-AUC/PR-AUC but the relation is very modest.
In general, it seems important that the RECols are properly specified and a good approximation to the underlying hypothesis while avoiding over-fitting.

(d) How should RECols be calculated and pre-processed? In addition to the choice of RECol algorithm one has many options to calculate and pre-process the created RECols. In terms of calculating we have tried mean squared errors and mean absolute deviations. We advise to try out both metrics because depending on the dataset and algorithm there were performance differences between the two methods. In principle one could think of other metrics too, like exponential distances. After having calculated the RECols it is important to scale them, for example by using a min-max scaler. Clipping RECol values or dropping RECols with high or low R-squared values have no big effects on outcomes, but might be relevant depending on the dataset at hand.

Apart from an evaluation on the aforementioned public datasets, we also successfully applied our approach to financial data using German Bundesbank internal datasets such as the Investment Funds Statistics (IFS) \cite{ifs}.

\subsection{Parameter Recommendations}
Based on the 88,693 experiments we ran to evaluate how to generate RECols across 10 standard datasets and 11 standard OD approaches for this study, we would now like to share some of our observations for practical use-cases, in which one might not want to run an equally comprehensive study.

In general, we would recommend to start with a simple random forest regressor as supervised algorithm to create the RECols based on the MSE metric.
This is due to our observation that this combination seems to be quite robust, often among the top performers and additionally has the advantage of only few parameters to tune. Wrt.\ scaling of input features and clipping the RECols the results are less conclusive, so we would recommend to experiment depending on the use case. Wrt.\ dropping RECols depending on $R^2$, we in did not observe a clear pattern, but the differences were negligible in most cases.
In case of many dimensions, we would hence suggest to drop the most uninformative RECols based on $R^2$ to reduce the load of downstream components.

\section{Conclusion and Future Work}\label{sec:conclusion}
In this paper we introduced our RECol pre-processing approach.
We showed how to calculate RECols and added them to the outlier detection algorithm as additional features.
We find that adding RECols can often significantly improve model performance of unsupervised outlier detection models in terms of the ROC-AUC and PR-AUC metric and using them does rarely harm model performance.
Based on more than 88k experiments, we also provide parameter recommendations to quickly try our approach in practical use-cases.
Further, simply fusing RECols to an outlier score (RECol-OD) delivers a surprisingly simple algorithm that outperforms standard algorithms in many cases.

Our results also open up many further avenues for future research.
So far we have not optimized hyper-parameters of the OD algorithms when using RECols.
This could further improve our results.
In addition one can experiment further in how to add RECols.
Maybe aggregating RECols to fewer columns based on dimensionality reduction techniques might be helpful to condense the information in the RECols in fewer columns that need to be added for training.
This is especially interesting as doubling the space of columns might be problematic in high-dimensional data.
Next, we have only applied a very simple fusion approach for the RECol-OD algorithm.
One can think of alternative and more complex approaches to fuse RECols that might improve the performance of the RECol-OD algorithm.
Finally, applying our approach to additional datasets, combining it with auto-encoders and also in real world applications might help to strengthen our findings further.

\bibliography{recol}

\begin{thebibliography}{23}
\providecommand{\natexlab}[1]{#1}
\providecommand{\url}[1]{\texttt{#1}}
\expandafter\ifx\csname urlstyle\endcsname\relax
  \providecommand{\doi}[1]{doi: #1}\else
  \providecommand{\doi}{doi: \begingroup \urlstyle{rm}\Url}\fi

\bibitem[Amarbayasgalan et~al.(2019)Amarbayasgalan, Park, Lee, and
  Ryu]{10.1371/journal.pone.0225991}
Amarbayasgalan, T., Park, K.~H., Lee, J.~Y., and Ryu, K.~H.
\newblock Reconstruction error based deep neural networks for coronary heart
  disease risk prediction.
\newblock \emph{PLOS ONE}, 14\penalty0 (12):\penalty0 1--17, 12 2019.
\newblock \doi{10.1371/journal.pone.0225991}.
\newblock URL \url{https://doi.org/10.1371/journal.pone.0225991}.

\bibitem[Amer \& Goldstein(2012)Amer and Goldstein]{ldcof}
Amer, M. and Goldstein, M.
\newblock Nearest-neighbor and clustering based anomaly detection algorithms
  for rapidminer.
\newblock In Fischer, S. and Mierswa, I. (eds.), \emph{Proceedings of the 3rd
  RapidMiner Community Meeting and Conferernce (RCOMM 2012). RapidMiner
  Community Meeting and Conference (RCOMM-2012), August 28-31, Budapest,
  Hungary}, pp.\  1--12. Shaker Verlag GmbH, 8 2012.
\newblock ISBN 978-3-8440-0995-8.

\bibitem[Amer et~al.(2013)Amer, Goldstein, and Abdennadher]{uocsvm}
Amer, M., Goldstein, M., and Abdennadher, S.
\newblock Enhancing one-class support vector machines for unsupervised anomaly
  detection.
\newblock In \emph{Proceedings of the ACM SIGKDD Workshop on Outlier Detection
  and Description}, ODD '13, pp.\  8–15, New York, NY, USA, 2013. Association
  for Computing Machinery.
\newblock ISBN 9781450323352.
\newblock \doi{10.1145/2500853.2500857}.
\newblock URL \url{https://doi.org/10.1145/2500853.2500857}.

\bibitem[Bache \& Lichman(2013)Bache and Lichman]{Bache+Lichman:2013}
Bache, K. and Lichman, M.
\newblock {UCI} machine learning repository, 2013.
\newblock URL \url{http://archive.ics.uci.edu/ml}.

\bibitem[Blaschke \& Haupenthal(2020)Blaschke and Haupenthal]{ifs}
Blaschke, J. and Haupenthal, H.
\newblock Investment funds statistics base, data report 2020-05.
\newblock \emph{Deutsche Bundesbank, Research Data and Service Centre}, 2020.

\bibitem[Breunig et~al.(2000)Breunig, Kriegel, Ng, and Sander]{lof}
Breunig, M.~M., Kriegel, H.-P., Ng, R.~T., and Sander, J.
\newblock Lof: Identifying density-based local outliers.
\newblock \emph{SIGMOD Rec.}, 29\penalty0 (2):\penalty0 93–104, May 2000.
\newblock ISSN 0163-5808.
\newblock \doi{10.1145/335191.335388}.
\newblock URL \url{https://doi.org/10.1145/335191.335388}.

\bibitem[Chalapathy \& Chawla(2019)Chalapathy and
  Chawla]{DBLP:journals/corr/abs-1901-03407}
Chalapathy, R. and Chawla, S.
\newblock Deep learning for anomaly detection: {A} survey.
\newblock \emph{CoRR}, abs/1901.03407, 2019.
\newblock URL \url{http://arxiv.org/abs/1901.03407}.

\bibitem[Chen et~al.(2009)Chen, Martin, and Montague]{CHEN20093706}
Chen, T., Martin, E., and Montague, G.
\newblock Robust probabilistic pca with missing data and contribution analysis
  for outlier detection.
\newblock \emph{Computational Statistics and Data Analysis}, 53\penalty0
  (10):\penalty0 3706 -- 3716, 2009.
\newblock ISSN 0167-9473.
\newblock \doi{https://doi.org/10.1016/j.csda.2009.03.014}.
\newblock URL
  \url{http://www.sciencedirect.com/science/article/pii/S0167947309001248}.

\bibitem[Geusebroek et~al.(2005)Geusebroek, Burghouts, and Smeulders]{aloi}
Geusebroek, J.-M., Burghouts, G.~J., and Smeulders, A. W.~M.
\newblock The amsterdam library of object images.
\newblock \emph{Int. J. Comput. Vision}, 61\penalty0 (1):\penalty0 103–112,
  January 2005.
\newblock ISSN 0920-5691.
\newblock \doi{10.1023/B:VISI.0000042993.50813.60}.
\newblock URL \url{https://doi.org/10.1023/B:VISI.0000042993.50813.60}.

\bibitem[Gogoi et~al.(2011)Gogoi, Bhattacharyya, Borah, and Kalita]{8130440}
Gogoi, P., Bhattacharyya, D., Borah, B., and Kalita, J.~K.
\newblock {A Survey of Outlier Detection Methods in Network Anomaly
  Identification}.
\newblock \emph{The Computer Journal}, 54\penalty0 (4):\penalty0 570--588, 03
  2011.
\newblock ISSN 0010-4620.
\newblock \doi{10.1093/comjnl/bxr026}.
\newblock URL \url{https://doi.org/10.1093/comjnl/bxr026}.

\bibitem[Goldstein \& Dengel(2012)Goldstein and Dengel]{hbos}
Goldstein, M. and Dengel, A.
\newblock Histogram-based outlier score (hbos): A fast unsupervised anomaly
  detection algorithm.
\newblock In Wölfl, S. (ed.), \emph{KI-2012: Poster and Demo Track. German
  Conference on Artificial Intelligence (KI-2012), 35th, September 24-27,
  Saarbrücken, Germany}, pp.\  59--63. Online, 9 2012.

\bibitem[Goldstein \& Uchida(2016)Goldstein and Uchida]{goldstein}
Goldstein, M. and Uchida, S.
\newblock A comparative evaluation of unsupervised anomaly detection algorithms
  for multivariate data.
\newblock \emph{PLOS ONE}, 11\penalty0 (4):\penalty0 1--31, 04 2016.
\newblock \doi{10.1371/journal.pone.0152173}.
\newblock URL \url{https://doi.org/10.1371/journal.pone.0152173}.

\bibitem[{Gong} et~al.(2019){Gong}, {Liu}, {Le}, {Saha}, {Reda Mansour},
  {Venkatesh}, and {van den Hengel}]{2019arXiv190402639G}
{Gong}, D., {Liu}, L., {Le}, V., {Saha}, B., {Reda Mansour}, M., {Venkatesh},
  S., and {van den Hengel}, A.
\newblock {Memorizing Normality to Detect Anomaly: Memory-augmented Deep
  Autoencoder for Unsupervised Anomaly Detection}.
\newblock \emph{arXiv e-prints}, art. arXiv:1904.02639, April 2019.

\bibitem[He et~al.(2003)He, Xu, and Deng]{cblof}
He, Z., Xu, X., and Deng, S.
\newblock Discovering cluster-based local outliers.
\newblock \emph{Pattern Recogn. Lett.}, 24\penalty0 (9–10):\penalty0
  1641–1650, June 2003.
\newblock ISSN 0167-8655.
\newblock \doi{10.1016/S0167-8655(03)00003-5}.
\newblock URL \url{https://doi.org/10.1016/S0167-8655(03)00003-5}.

\bibitem[Jablonski et~al.(2015)Jablonski, Bihl, and Jr.]{7098354}
Jablonski, J.~A., Bihl, T.~J., and Jr., K. W.~B.
\newblock Principal component reconstruction error for hyperspectral anomaly
  detection.
\newblock \emph{{IEEE} Geosci. Remote. Sens. Lett.}, 12\penalty0 (8):\penalty0
  1725--1729, 2015.
\newblock \doi{10.1109/LGRS.2015.2421813}.
\newblock URL \url{https://doi.org/10.1109/LGRS.2015.2421813}.

\bibitem[Kriegel et~al.(2009)Kriegel, Kr\"{o}ger, Schubert, and Zimek]{loop}
Kriegel, H.-P., Kr\"{o}ger, P., Schubert, E., and Zimek, A.
\newblock Loop: Local outlier probabilities.
\newblock In \emph{Proceedings of the 18th ACM Conference on Information and
  Knowledge Management}, CIKM '09, pp.\  1649–1652, New York, NY, USA, 2009.
  Association for Computing Machinery.
\newblock ISBN 9781605585123.
\newblock \doi{10.1145/1645953.1646195}.
\newblock URL \url{https://doi.org/10.1145/1645953.1646195}.

\bibitem[Liu et~al.(2008)Liu, Ting, and Zhou]{iso}
Liu, F.~T., Ting, K.~M., and Zhou, Z.-H.
\newblock Isolation forest.
\newblock In \emph{Proceedings of the 2008 Eighth IEEE International Conference
  on Data Mining}, ICDM '08, pp.\  413–422, USA, 2008. IEEE Computer Society.
\newblock ISBN 9780769535029.
\newblock \doi{10.1109/ICDM.2008.17}.
\newblock URL \url{https://doi.org/10.1109/ICDM.2008.17}.

\bibitem[Micenková et~al.(2014)Micenková, McWilliams, and
  Assent]{Micenkov2014LearningOE}
Micenková, B., McWilliams, B., and Assent, I.
\newblock Learning outlier ensembles: The best of both worlds–supervised and
  unsupervised.
\newblock \emph{ACM SIGKDD 2014 Workshop ODD}, 2014.

\bibitem[Ramaswamy et~al.(2000)Ramaswamy, Rastogi, and Shim]{kthnn}
Ramaswamy, S., Rastogi, R., and Shim, K.
\newblock Efficient algorithms for mining outliers from large data sets.
\newblock \emph{SIGMOD Rec.}, 29\penalty0 (2):\penalty0 427–438, May 2000.
\newblock ISSN 0163-5808.
\newblock \doi{10.1145/335191.335437}.
\newblock URL \url{https://doi.org/10.1145/335191.335437}.

\bibitem[Sakurada \& Yairi(2014)Sakurada and Yairi]{Sakurada}
Sakurada, M. and Yairi, T.
\newblock Anomaly detection using autoencoders with nonlinear dimensionality
  reduction.
\newblock In \emph{Proceedings of the MLSDA 2014 2nd Workshop on Machine
  Learning for Sensory Data Analysis}, MLSDA'14, pp.\  4–11, New York, NY,
  USA, 2014. Association for Computing Machinery.
\newblock ISBN 9781450331593.
\newblock \doi{10.1145/2689746.2689747}.
\newblock URL \url{https://doi.org/10.1145/2689746.2689747}.

\bibitem[Sch\"{o}lkopf et~al.(2001)Sch\"{o}lkopf, Platt, Shawe-Taylor, Smola,
  and Williamson]{ocsvm}
Sch\"{o}lkopf, B., Platt, J.~C., Shawe-Taylor, J.~C., Smola, A.~J., and
  Williamson, R.~C.
\newblock Estimating the support of a high-dimensional distribution.
\newblock \emph{Neural Comput.}, 13\penalty0 (7):\penalty0 1443–1471, July
  2001.
\newblock ISSN 0899-7667.
\newblock \doi{10.1162/089976601750264965}.
\newblock URL \url{https://doi.org/10.1162/089976601750264965}.

\bibitem[Xia et~al.(2015)Xia, Cao, Wen, Hua, and Sun]{Xia}
Xia, Y., Cao, X., Wen, F., Hua, G., and Sun, J.
\newblock Learning discriminative reconstructions for unsupervised outlier
  removal.
\newblock In \emph{2015 {IEEE} International Conference on Computer Vision,
  {ICCV} 2015, Santiago, Chile, December 7-13, 2015}, pp.\  1511--1519. {IEEE}
  Computer Society, 2015.
\newblock \doi{10.1109/ICCV.2015.177}.
\newblock URL \url{https://doi.org/10.1109/ICCV.2015.177}.

\bibitem[Zhou \& Paffenroth(2017)Zhou and Paffenroth]{Zhou}
Zhou, C. and Paffenroth, R.~C.
\newblock Anomaly detection with robust deep autoencoders.
\newblock In \emph{Proceedings of the 23rd ACM SIGKDD International Conference
  on Knowledge Discovery and Data Mining}, KDD '17, pp.\  665–674, New York,
  NY, USA, 2017. Association for Computing Machinery.
\newblock ISBN 9781450348874.
\newblock \doi{10.1145/3097983.3098052}.
\newblock URL \url{https://doi.org/10.1145/3097983.3098052}.

\end{thebibliography}
\bibliographystyle{icml2021}

\end{document}